# A Multimodal Biometric System Using Linear Discriminant Analysis For Improved Performance

Aamir Khan[1], Muhammad Farhan[2], Aasim Khurshid[3] and Adeel Akram[4]

[1,4] Electrical Engineering Department, COMSATS Institute of IT Wah Campus
Wah Cantt., Punjab 47040, Pakistan

[2] Institute of Engineering and Computing Sciences, University of Science and Technology Bannu
Bannu, KPK 28100, Pakistan

[3] Software Engineering Department, University of Engineering and Technology Taxila
Taxila, Punjab 47080, Pakistan

**Abstract**

Essentially a biometric system is a pattern recognition system which recognizes a user by determining the authenticity of a specific anatomical or behavioral characteristic possessed by the user. With the ever increasing integration of computers and Internet into daily life style, it has become necessary to protect sensitive and personal data. This paper proposes a multimodal biometric system which incorporates more than one biometric trait to attain higher security and to handle failure to enroll situations for some users. This paper is aimed at investigating a multimodal biometric identity system using Linear Discriminant Analysis as backbone to both facial and speech recognition and implementing such system in real-time using SignalWAVE.

***Keywords:*** *Pattern Recognition, PCA, LDA, KNN, MFCCs, SVM, Speech/Visual Authentication.*

## 1. Introduction

The system automatically recognizes or identifies the user based on voice and facial information. The general audio-visual recognition system is shown in Figure 1.1.

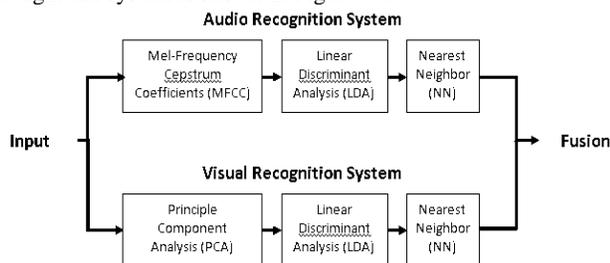

**Figure** 1.1 Audio-Visual User Recognition Systems

The system takes in the inputs and passes through two modules built in it namely visual recognition system and audio recognition system. A biometric template or reference for a user is first captured in order to enroll in the system. The said reference is securely stored in a central database or it could be smart card issued to the user. The template is used for matching when an individual needs to be identified. This multimodal biometric system can operate either in verification (authentication) or an identification mode based on context. Performance of the LDA on two biometric traits that the system uses is briefly discussed.

**Visual Recognition System** makes an attempt to match the facial features of a user to its template in the data base. It uses Principal Component Analysis, Linear Discriminant Analysis and K nearest neighbor.
The Principal Component Analysis reduces the large dimensionality of the data space (observed variables) to the smaller intrinsic dimensionality of feature space (independent variables), which are needed to describe the data economically.
Unlike PCA, LDA explicitly attempts to model the difference between the classes of data [6] and factor analysis builds the feature combinations based on differences rather than similarities. LDA is also different from factor analysis in that it is not an interdependence technique: a distinction between independent variables and dependent variables (also called criterion variables) must be made. Therefore LDA significantly reduces the dimensionality of the data obtained after PCA and with the KNN classifier produces significant improvement in performance.

**Audio Recognition System** takes into account the fact that Performance of a speech recognition system is affected considerably by the choice of features for most of the applications. Raw data obtained from speech recording can't be used to directly train the recognizer as for the same phonemes do not necessarily have same sample values. Here we proposed a system which increases the efficiency of support vector machines by apply this technique after LDA implementation.
Block diagram of Mel-frequency cepstral coefficients (MFCCs) processor is shown in Figure 1.2.

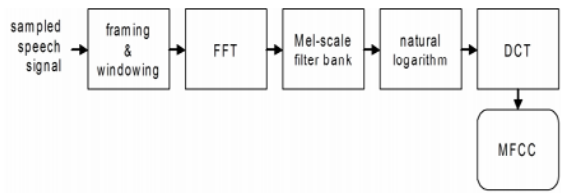

**Figure** 1.1 Block diagram of the MFCC processor

Mel-frequency cepstral are essentially the representation of power spectrum of sound signals which are made up from coefficients, based on a linear cosine transform of a log power spectrum on a nonlinear mel scale of frequency.

Use of LDA after mfcc drastically reduces the dimension of features as LDA finds optimal transformation matrix which preserve most of the information and the same can be used to discriminate between the different classes. Further it is necessary to use a label of the recorded data as to associate each speech segment with a label.

Support Vector Machine i.e. SVMs were introduced by Boser, Guyon and Vapnik in COLT-92 [28-30], works on principle of based on some previous training through inputs, using supervised learning techniques to classify data [31]. Using machine learning theory to boost higher accuracy and avoid the data to be over-fit automatically, SVM is classifier. SVM has been tested for hand writing, face recognition in general pattern classification and regression based applications.

After 100% successful simulation results of the multimodal system proposed, same system was implemented onto the Field Programmable Gate Array (FPGA) using SignalWave.

## 2. Methodology

For the visual recognition systems following steps are taken. The resulting vector from 2 D should present as:
$$x^i = [x_1^i \ldots x_N^i]^T$$

The mean image is a column vector
$$\bar{x}^i = x^i - m$$

Where
$$m = \frac{1}{p}\sum_{i=1}^{p} x^i$$

Center data
$$\bar{X} = \lfloor \bar{x}^1 | \bar{x}^2 | \ldots | \bar{x}^p | \rfloor$$

Covariance matrix
$$\Omega = \overline{XX}^T$$

Sorting the order of Eigenvectors
$$V = \lfloor V_1 | V_2 | \ldots | V_p \rfloor$$

Projection of the image
$$\tilde{x}^i = V^T \bar{x}^i$$

Identifying the new images

$$\bar{y}^i = y^i - m$$

and

$$\tilde{y}^i = V^T \bar{y}^i$$

To reduce the dimensionality LDA is performaed by calculating the Class Matrix, Between Class Matrix, Generalized Eigenvalue and Eigenvector, sorting the order of eigenvector and projecting training images onto the fisher basis vectors.

$$S_i = \sum_{x \in X_i}(x - m_i)(x - m_i)^T$$
$$S_W = \sum_{i=1}^{C} S_i$$

Where C is the number of classes
$$S_B = \sum_{i=1}^{C} n_i (x - m_i)(x - m_i)^T$$

Where $n_i$ is the number of image in the *ith* class, *m* is the total mean of all the training images.

Euclidean distance is used to compute the distance. The mathematic formula of Euclidean distance is presented as:

$$D(x_j, y_j) = \sqrt{\sum_{j=1}^{n}(x_j - y_j)^2}$$

By comparing the output of this operation to a set of sample result we identified five clients in the data set.

For the Audio Recognition System, The result of windowing is the signal

$$y_l(n) = x_l(n)h(n), \quad 0 \leq n \leq N - 1$$

Typically the Hamming window is used, which has the form:

$$h(n) = 0.54 - 0.46\cos\left(\frac{2\pi n}{N-1}\right), \quad 0 \leq n \leq N - 1$$

Fast Fourier Transform converts each frame of *N* samples from the time domain into the frequency domain.

$$X_k = \sum_{n=0}^{N-1} x_n e^{-j2\pi kn/N}, \quad k = 0,1,2,\ldots, N-1$$

The result after this step is often referred to as spectrum. That filter bank has a triangular bandpass frequency response, and the spacing as well as the bandwidth is determined by a constant mel frequency interval. The number of mel spectrum coefficients, *K*, is typically chosen as 20.

*converting them to the time domain using the Discrete Cosine Transform (DCT).*

$$\tilde{c}_n = \sum_{k=1}^{K}(\log S_k)\cos\left[n\left(k - \frac{1}{2}\right)\frac{\pi}{K}\right], \quad n=0,1,\ldots,K-1$$

Where n=1,2,….K

Note that we exclude the first component, $\tilde{c}_0$, from the DCT since it represents the mean value of the input signal, which carried little speaker specific information.

In order to reduce the large dimension of the features, most discriminating speech features are further acquired by using the LDA by calculating the Class Matrix, Between Class Matrix, Generalized Eigenvalue and Eigenvector, sorting the order of eigenvector and projecting training records onto the fisher basis vectors.

$$S_{\bar{i}} = \sum_{x \in X_i}(x - m_i)(x - m_i)^T$$
$$S_W = \sum_{i=1}^{C} S_{\bar{i}}$$

Where C is the number of classes

$$S_B = \sum_{i=1}^{C} n_i (x - m_i)(x - m_i)^T$$

Where $n_i$ is the number of feature data in the *ith* class, *m* is the total mean of all the training recorded data.

For linear SVM, separating data with a hyper plane and extending to non-linear boundaries by using kernel trick [18, 12] and correct classification of all data is mathematically

i)  $wx_i + b \geq 1$, if $y_i = +1$

ii) $wx_i + b \leq 1$, if $y_i = -1$

iii) $y_i (w_i + b) \geq 1$, for all i

Where w is a weight vector. Depending upon the quality of the training data to be good and every test vector in the range of radius r from the training vector, the hyper plane selected is considered at the farthest possible region from the data. Linear constraints are required to optimize the quadratic function

$$f(\mathbf{x}) = \Sigma a_i\, y_i\, \mathbf{x}_i\, * \mathbf{x} + b$$

For data with linear attribute, a separating hyper plane suffice to divide the data, however, in most cases the data is non linear and inseparable. Kernels use the non-linearly mapped data into a space of high dimensions and distinguish the linear data [31]. For example in [36]

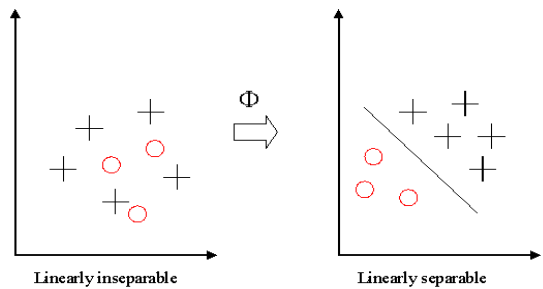

**Figure** 2.1
Kernel defines the mapping as

$$k(x, y) = \emptyset(x).\emptyset(y)$$

After transforming data into feature space, similarity measure can be defined on basis of dot product. A good choice of feature space makes the pattern recognition easier [31].

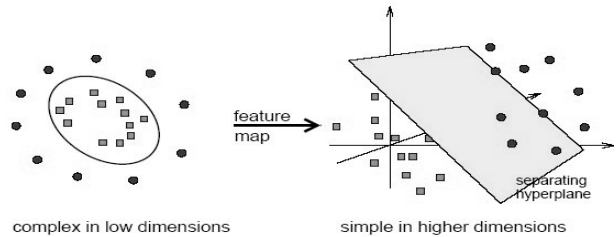

**Figure** 3.2 Feature Space Representation

SVM are used as classifiers but recently have been used for regression purposes [34].

For implementation of the system on FPGA using SignalWave, most of the blocks were taken from the workspace especially in the PCA and LDA Simulink blockset. The Nearest neighbor simulink block was designed entirely from the program. As Simulink is an integral part of MATLAB, it is easy to switch back and forth during the design though the logical interconnects from the program is of paramount importance. When performing real-time operation using this design all the results obtained were correct.

## 3. Results

The outcome of the visual recognition system is shown in the Figure 3.1 and Figure 3.2.

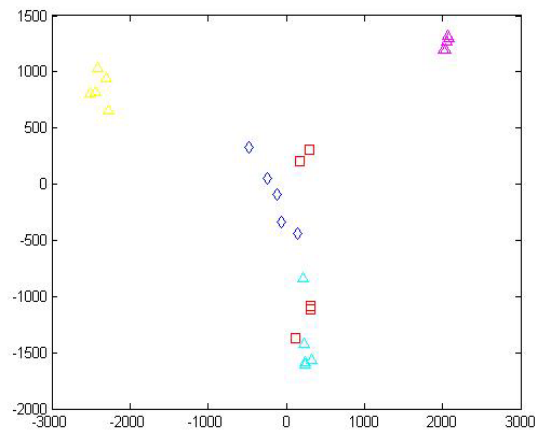

**Figure** 3.1 PCA plot showing 5 clients in no particular order.

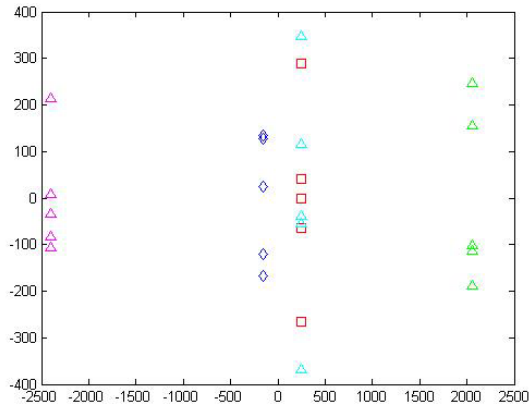

**Figure** 3.2 LDA plot.

The result for the sound file chosen from the testing sound and it was the 5th sound file as there are two test sound files for each word. Hence it is shown that SVM efficiency is reduced when the dimensionality of the training data is increased. It is shown that if LDA and SVM are both used together produce very good results as shown in Figure.1. And SVM to audio data before LDA and LDA projection train are shown in Figure. 3.3 and Figure. 3.4 Respectively.

```
>> svmtrain(classlabel, ldaprojtrainingimg','-g 2 -c 10 -v 10')
Cross Validation Accuracy = 100%

ans =

    100

>>
```

**Figure. 3.3:** SVM after LDA Cross-validation

```
>> svmtrain(classlabel, audiodata','-g 2 -c 10 -v 10')
Cross Validation Accuracy = 60%

ans =

    60

>>
```

**Figure. 3.4:** SVM to Audio data before LDA

```
>> model= svmtrain(classlabel, ldaprojtrainingimg','-g 2 -c 10')

model =

    Parameters: [5x1 double]
      nr_class: 5
       totalSV: 5
           rho: [10x1 double]
         Label: [5x1 double]
         ProbA: []
         ProbB: []
           nSV: [5x1 double]
      sv_coef: [5x4 double]
           SVs: [5x5 double]

>> predict= svmpredict(classlabel2,ldaprojtestimage,model)
Accuracy = 40% (2/5) (classification)

predict =

    1
    1
    5
    5
    5

>>
```

**Figure. 3.5:** LDA Projection train

After making each block in Simulink, constant referral was made to the workpsace in order to match the output from simulink blocks to the output from the code. The final result which was obtained at the display block is shown below.

For K=2

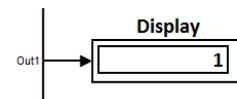

**Figure** 3.6 Result for k=2 (test database)

For K=5:

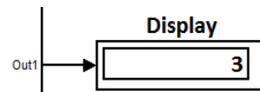

**Figure** 3.7 Result for k=5 (test database)

For K=10:

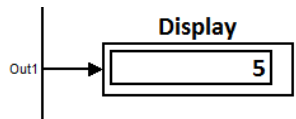

**Figure** 3.9 Result for k=10 (test database)

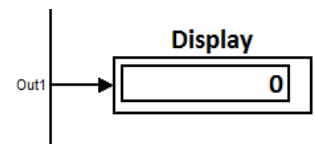

**Figure** 3.10 Result for k=2 (fraud database)

## 4. Conclusions

The multimodal biometric system proposed in this paper is tested for 100% accuracy. Also it is reported that the use of LDA highly decreases the complexity in pattern recognition. It was noticed that SVM alone for speech recognition produces poor results and the success rate is 60% while when it is used together with LDA the success rate was noted for 100% and improvement rate of 40% is achieved. Furthermore, it is recommended to experiment HLDA with support vector machine (SVM) for speech recognition.

**Aamir Khan** has received M.Sc. degree in Electronic Communications and Computer Engineering from University of Nottingham Malaysia Campus in 2011. He is working at CIIT Wah Campus as lecturer in Electrical Engineering Department. Research interests include embedded systems, intelligent systems, pattern recognition and optical communications.

**Muhammad Farhan** has received M.Sc. degree in Electronic Communications and Computer Engineering from University of Nottingham Malaysia Campus in 2011. He is working at University of Science and Technology Bannuuas lecturer in Institute of Engineering and Computing Sciences. Research interests include speech processing, intelligent systems and pattern recognition.

**Aasim Khurshid** is MSc Software Engineering from University of Science and Technology, Taxila, Pakistan and is working at Government Post Graduate College, Mandian, Pakistan. Research interests include operating systems, pattern recognition.

**Adeel Akram** has received M.Sc. degree in Computer Engineering from National University of Science and Technology Pakistan. He is working at CIIT Wah Campus as lecturer in Electrical Engineering Department. Research interests include embedded systems, reconfigurable computing, pattern recognition and computer architecture.